\documentclass{article}

\usepackage{microtype}
\usepackage{graphicx}
\usepackage{subcaption}
\usepackage{booktabs} % for professional tables

\usepackage{amsmath,amssymb,amsthm,bbm,mdframed,relsize}
\usepackage{hyperref}

\usepackage[accepted]{icml2019}

\icmltitlerunning{Locally Private Bayesian Inference for Count Models}

\newtheorem{theorem}{Theorem}
\newtheorem{definition}{Definition}
\usepackage[]{color-edits}
\addauthor{sw}{blue}

\newcommand{\subs}{\mathbf{i}}
\newcommand{\ys}{y_{\subs}}
\newcommand{\ysp}{\tilde{y}_{\subs}^{\mathsmaller{(+)}}}
\newcommand{\yspm}{\tilde{y}_{\subs}^{\mathsmaller{(\pm)}}}
\newcommand{\mus}{\mu_{\subs}}

\newcommand{\lambdap}{\lambda^{\mathsmaller{(+)}}_{\subs}}
\newcommand{\lambdam}{\lambda^{\mathsmaller{(-)}}_{\subs}}
\newcommand{\gp}{g^{\mathsmaller{(+)}}_{\subs}}
\newcommand{\gm}{g^{\mathsmaller{(-)}}_{\subs}}

\newcommand{\given}{\,|\,}
\newcommand{\teq}{\!=\!}
\newcommand{\tp}{\!+\!}
\newcommand{\tm}{\!-\!}
\newcommand{\tequiv}{\!\equiv\!}
\newcommand{\ttimes}{\!\times\!}
\newcommand{\tsim}{\!\sim\!}
\newcommand{\tin}{\!\in\!}

\newcommand{\Pois}{\textrm{Pois}}
\newcommand{\compcond}[1]{\big(#1\given-\big)}

\newcommand{\naive}{na\"ive }

\begin{document}

\twocolumn[
\icmltitle{Locally Private Bayesian Inference for Count Models}



\icmlsetsymbol{equal}{*}

\begin{icmlauthorlist}
\icmlauthor{Aaron Schein}{umass}
\icmlauthor{Zhiwei Steven Wu}{msr}
\icmlauthor{Alexandra Schofield}{cornell}
\icmlauthor{Mingyuan Zhou}{ut}
\icmlauthor{Hanna Wallach}{msr}
\end{icmlauthorlist}

\icmlaffiliation{umass}{University of Massachusetts Amherst}
\icmlaffiliation{msr}{Microsoft Research, New York}
\icmlaffiliation{ut}{University of Texas at Austin}
\icmlaffiliation{cornell}{Cornell University}

\icmlcorrespondingauthor{Aaron Schein}{aschein@cs.umass.edu}
% \icmlcorrespondingauthor{Zhiwei Steven Wu}{zsw@umn.edu}
% \icmlcorrespondingauthor{Alexandra Schofield}{xanda@cs.cornell.edu}
% \icmlcorrespondingauthor{Mingyuan Zhou}{mingyuan.zhou@mccombs.utexas.edu}
% \icmlcorrespondingauthor{Hanna Wallach}{hanna@dirichlet.net}

%\icmlkeywords{Machine Learning, ICML}

\vskip 0.2in
]



\printAffiliationsAndNotice{}  % leave blank if no need to mention equal contribution

\begin{abstract}
We present a general method for privacy-preserving Bayesian inference in Poisson factorization, a broad class of models that includes some of the most widely used models in the social sciences. Our method satisfies limited precision local privacy, a generalization of local differential privacy, which we introduce to formulate privacy guarantees appropriate for sparse count data. We develop an MCMC algorithm that approximates the locally private posterior over model parameters given data that has been locally privatized by the geometric mechanism~\citep{ghosh2012universally}. Our solution is based on two insights:~1) a novel reinterpretation of the geometric mechanism in terms of the Skellam distribution~\citep{skellam1946frequency} and 2) a general theorem that relates the Skellam to the Bessel distribution~\citep{yuan2000bessel}. We demonstrate our method in two case studies on real-world email data in which we show that our method consistently outperforms the commonly-used \naive approach, obtaining higher quality topics in text and more accurate link prediction in networks. On some tasks, our privacy-preserving method even outperforms non-private inference which conditions on the true data.~\looseness=-1
\end{abstract}

\section{Introduction}
\label{sec:intro}

Data from social processes often take the form of discrete observations (e.g., edges in a social network, word tokens in an email) and these observations often contain sensitive information. As more aspects of social interaction are digitally recorded, the opportunities for social scientific insights grow; however, so too does the risk of unacceptable privacy violations.  As a result, there is a growing need to develop privacy-preserving data analysis methods. In practice, social scientists will be more likely to adopt these methods if doing so entails minimal change to their current methodology.~\looseness=-1

Toward that end, we present a method for privacy-preserving Bayesian inference in Poisson factorization~\citep{titsias2008infinite,cemgil2009bayesian,zhou2012augment,gopalan2013efficient,paisley2014bayesian}, a broad class of models for learning latent structure from discrete data.  This class contains many of the most widely used models in the social sciences, including topic models for text corpora~\citep{blei2003latent,buntine2004applying,canny2004gap}, population models for genetic data~\citep{pritchard2000inference}, stochastic block models for social networks~\citep{ball2011efficient,gopalan2013efficient,zhou2015infinite}, and tensor factorization for dyadic data~\citep{welling2001positive,chi2012tensors,schmidt2013nonparametric,schein2015bayesian,schein2016bayesian}. It further includes deep hierarchical models~\citep{ranganath2015deep,zhou2015poisson}, dynamic models~\citep{charlin2015dynamic,acharya2015nonparametric,schein2016pgds}, and many others.

Our method applies principles of differential privacy~\citep{dwork2006calibrating} to observations locally before they are aggregated into a data set for analysis; this ensures that no single centralized server need ever store the non-privatized data, a condition that is often desirable. We introduce \textit{limited precision local privacy (LPLP)}---which generalizes the standard definition of local differential privacy---in order to formulate privacy guarantees that are natural for applications like topic modeling, in which observations (i.e., documents) are high-dimensional vectors of sparse counts. To privatize count data, our method applies the geometric mechanism~\citep{ghosh2012universally} which we prove is a mechanism for LPLP.~\looseness=-1

\begin{figure*}[t]
\centering
  \begin{subfigure}[t]{0.49\linewidth}
      \centering
      \includegraphics[width=1.\linewidth]{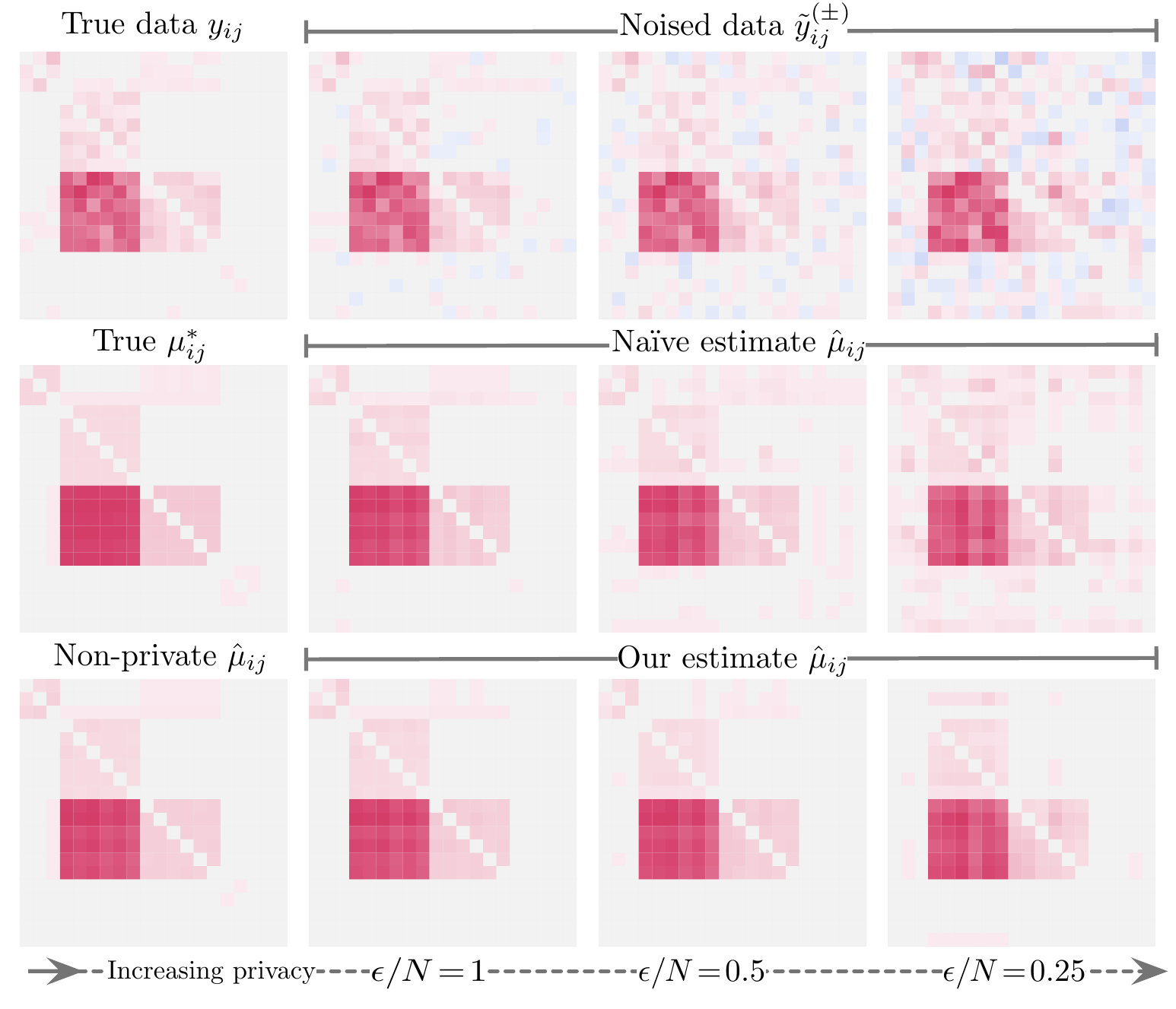}
  \end{subfigure}%
  ~
  \begin{subfigure}[t]{0.49\linewidth}
      \centering
      \includegraphics[width=1.\linewidth]{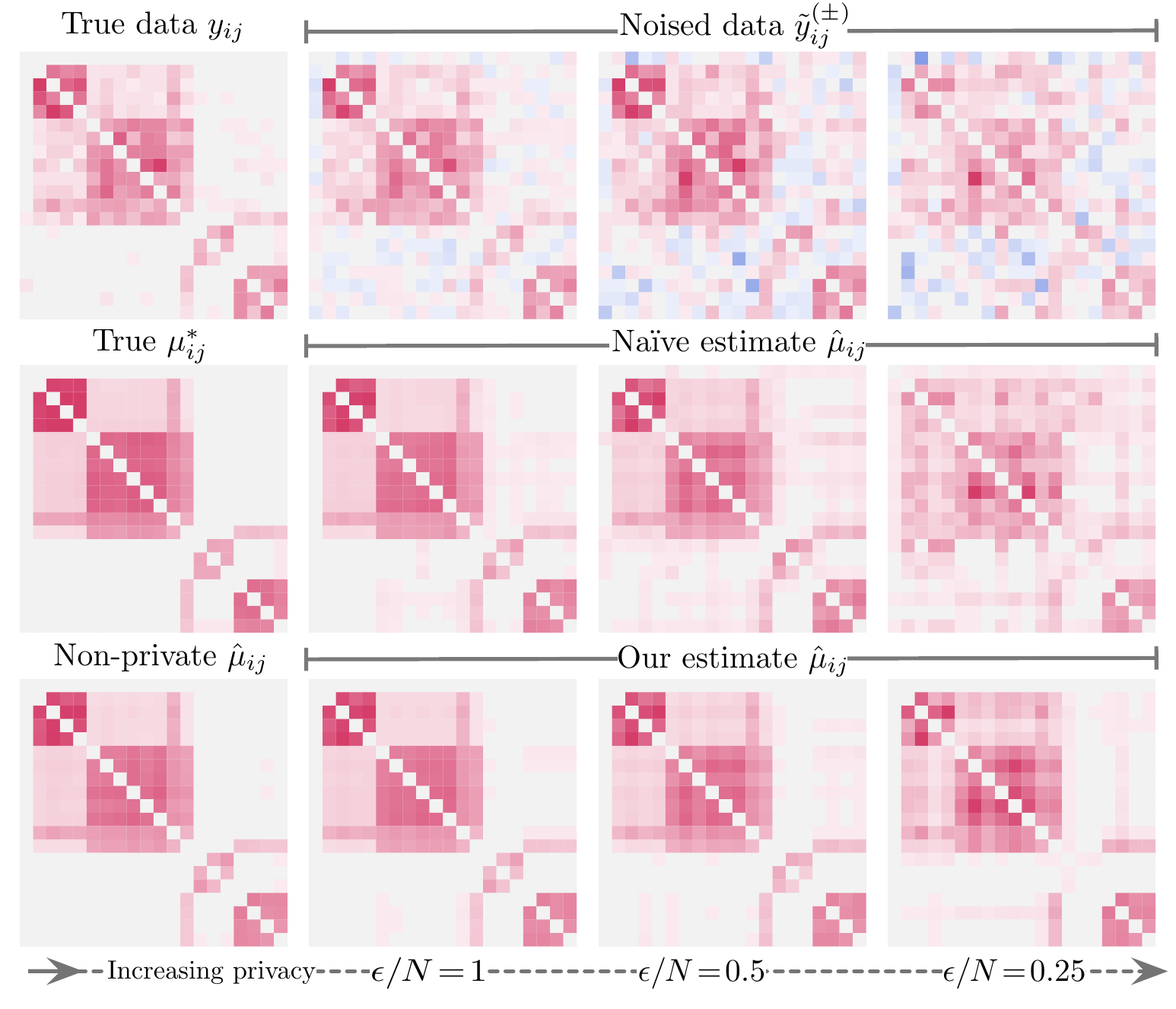}
      % \caption{\label{subfig:bar}}s
  \end{subfigure}%
\caption{\label{fig:synth-community} Block structure recovery: our method vs.
the \naive approach. We generated the non-privatized data synthetically. We then privatized the data using three different levels of noise. The top row depicts the data, using red to denote positive observations and blue to denote negative observations. As privacy increases, the \naive approach overfits the noise and fails to recover the true $\mu^{\star}_{ij}$ values, predicting high values even for sparse parts of the matrix. In contrast, our method recovers the latent structure, even at high noise levels.}
\vspace{-1em}
\end{figure*}

Under local privacy, a data analysis algorithm sees only privatized (i.e., noised) versions of data. A key statistical challenge is how to fit a model of the true data having only observed the privatized data. One option is a \naive approach, wherein inference proceeds as usual, treating the privatized data as if it were not privatized. In the context of maximum likelihood estimation, the \naive approach has been shown to exhibit pathologies when observations are discrete or count-valued; researchers have therefore advocated for treating the non-privatized observations as latent variables to be inferred~\citep{yang2012differential,karwa2014differentially,bernstein2017differentially}. We embrace this approach and extend it to Bayesian inference, where our goal is to approximate the \textit{locally private posterior} distribution over latent variables conditioned on the privatized data and randomized response method.~\looseness=-1

\pagebreak
Towards that goal, we introduce a novel Markov chain Monte Carlo (MCMC) algorithm that is asymptotically guaranteed to sample from the locally private posterior. Our algorithm iteratively re-samples values of the true underlying data from its inverse distribution conditioned on the model latent variables, and then vice versa; as a result our algorithm is modular, allowing a user to re-purpose an existing implementation of non-private MCMC which samples the model latent variables given (samples of) the true data.~\looseness=-1 

Our main technical contribution is the derivation of an analytically closed-form and computationally tractable scheme for ``inverting the noise process'' to sample values of the true data. Our solution relies two key insights: 1) a novel reinterpretation of the geometric mechanism in terms of the Skellam distribution~\citep{skellam1946frequency}, and 2) a theorem that relates the Skellam to the Bessel~\citep{yuan2000bessel} distribution. These insights are generally applicable and may be of independent interest in other contexts.~\looseness=-1

We present two case studies applying our method to 1) topic modeling for text corpora and, 2) overlapping community detection in social networks. We compare our method to the \naive approach over a range of privacy levels on synthetic data and real-world data from the Enron email corpus~\citep{klimt2004enron}. Our method consistently outperforms the \naive approach on a range of intrinsic and extrinsic evaluation metrics---notably, our method obtains more coherent topics from text and more accurate predictions of missing links in networks. We provide an illustrative example in figure~\ref{fig:synth-community}. Finally, on some tasks, our method even outperforms the non-private method which conditions on the true data. This finding suggests that Poisson factorization, as widely deployed in practice, is under-regularized and prompts future research into applying and understanding our method as a general method for robust count modeling.\looseness=-1

%%%%%%%%%%%%%%%%%%%%%%%%%%%%%%%%%%%%%%%%%%%
\section{Background and problem formulation}
\label{sec:bg}

\textbf{Differential privacy.} Differential privacy~\citep{dwork2006calibrating} is a rigorous privacy criterion that guarantees that no single observation in a data set will have a significant influence on the information obtained by analyzing that data set.\looseness=-1

\begin{definition}
\label{def:dp}
  A randomized algorithm $\mathcal{A}(\cdot)$ satisfies
  $\epsilon$-differential privacy if for all pairs of neighboring data
  sets $Y$ and $Y'$ that differ in only a single observation
  \begin{equation}
    P\left(\mathcal{A}(Y) \in \mathcal{S}\right) \leq e^\epsilon
      P\left(\mathcal{A}(Y') \in \mathcal{S}\right),
  \end{equation}
for all subsets $\mathcal{S}$ in the range of $\mathcal{A}(\cdot)$.
\end{definition}

\textbf{Local differential privacy.} We focus on local differential privacy,
which we refer to as local privacy. Under this criterion, the observations remain private from even the data analysis algorithm. The algorithm only sees privatized versions of the observations, often constructed by adding noise from specific distributions. The process of adding noise is known as randomized response---a reference to survey-sampling methods originally developed in the social sciences prior to the development of differential privacy~\cite{warner}. Satisfying this criterion means one never needs to aggregate the true data in a single location.\looseness=-1

\pagebreak
\begin{definition}
  \label{def:local_privacy}
A randomized response method $\mathcal{R}(\cdot)$ is $\epsilon$-private if for all pairs of observations $y, y' \in \mathcal{Y}$\looseness=-1
\begin{equation}
  \label{eqn:local_privacy}
    P\left(\mathcal{R}(y) \in \mathcal{S}\right) \leq e^\epsilon
      P\left(\mathcal{R}(y') \in \mathcal{S}\right),
        \end{equation}
for all subsets $\mathcal{S}$ in the range of $\mathcal{R}(\cdot)$. If a data analysis algorithm sees only the observations'
$\epsilon$-private responses, then the data analysis itself satisfies
$\epsilon$-local privacy.
\end{definition}

The meaning of ``observation'' in definitions~\ref{def:dp} and~\ref{def:local_privacy} varies across applications. In the context of topic modeling, an observation is an individual document $y \in \mathbbm{Z}_+^{V}$ such that each single entry $y_{v} \in \mathbbm{Z}_+$ is the count of word tokens of type $v$ in the document.  To guarantee local privacy, the randomized response method has to satisfy the condition in equation~\ref{eqn:local_privacy} for any pair of observations. This typically involves adding noise that scales with the maximum difference between any pair of observations defined as $N^{(\textrm{max})} = \max_{y, y'}\|y \tm y'\|_1$. When the observations are documents, $N^{(\textrm{max})}$ can be prohibitively large and the amount of noise will overwhelm the signal in the data. This motivates the following alternative formulation of privacy.~\looseness=-1

\textbf{Limited-precision local privacy.}
While standard local privacy requires that arbitrarily different observations become indistinguishable under the randomized response method, this guarantee may be unnecessarily strong in some settings. For instance, suppose a user would like to hide only the fact that their email contains a handful of vulgar curse words. Then it is sufficient to have a randomized response method which guarantees that any two similar emails---one containing the vulgar curse words and the same email without them---will be rendered indistinguishable after randomization. In other words, this only requires the randomized response method to render small neighborhoods of possible observations indistinguishable. To operationalize this kind of guarantee, we generalize definition~\ref{def:local_privacy} and define limited-precision local privacy (LPLP).~\looseness=-1

\begin{definition}
\label{def:lplp}
For any positive integer $N$, we say that a randomized response method
$\mathcal{R}(\cdot)$ is $(N, \epsilon)$-private if for all pairs of observations $y, y' \in \mathcal{Y}$ such that $\|y - y'\|_1 \leq N$
\begin{equation}
  \label{eqn:lp_local_privacy}
    P\left(\mathcal{R}(y) \in \mathcal{S}\right) \leq e^\epsilon
      P\left(\mathcal{R}(y') \in \mathcal{S}\right),
        \end{equation}
for all subsets $\mathcal{S}$ in the range of $\mathcal{R}(\cdot)$. If a data analysis algorithm sees only the observations' $(N,
\epsilon)$-private responses, then the data analysis itself satisfies
$(N, \epsilon)$-limited-precision local privacy. If $\|y\|_1 \leq N$
for all $y \in \mathcal{Y}$, then $(N, \epsilon)$-limited-precision local privacy implies $\epsilon$-local privacy.
\end{definition}

LPLP is the local privacy analog to limited-precision differential privacy, originally proposed by~\citet{flood} and subsequently used to privatize analyses of geographic location data~\citep{geo} and financial network data~\citep{dstress}. We emphasize that this is a \emph{strict generalization} of local privacy. A randomized response method that satisfies LPLP adds noise which scales as a function of $N$ and $\epsilon$---thus the same method may be interpreted as being $\epsilon$-private for a given setting of $N$ or $\epsilon'$-private for a different setting $N'$. In section~\ref{sec:poisson_factorization}, we describe the geometric mechanism~\citep{ghosh2012universally} and show how it satisfies LPLP.~\looseness=-1

\textbf{Differentially private Bayesian inference.} In Bayesian statistics, we begin with a probabilistic model $\mathcal{M}$ that relates observable variables $Y$ to latent variables $Z$ via a joint distribution $P_{\mathcal{M}}(Y, Z)$.  The goal of inference is then to compute the posterior distribution $P_{\mathcal{M}}(Z\,|\,Y)$ over the latent variables conditioned on observed values of $Y$.  The posterior is almost always analytically intractable and thus inference involves approximating it.  The two most common methods of approximate Bayesian inference are variational inference, wherein we fit the parameters of an approximating distribution $Q(Z\,|\,Y)$, and Markov chain Monte Carlo (MCMC), wherein we approximate the posterior with a finite set of samples $\{Z^{(s)}\}_{s=1}^S$ generated via a Markov chain whose stationary distribution is the exact posterior.~\looseness=-1

We can conceptualize Bayesian inference as a randomized algorithm $\mathcal{A}(\cdot)$ that returns an approximation to the posterior distribution $P_{\mathcal{M}}(Z\,|\,Y)$. In general $\mathcal{A}(\cdot)$ does not satisfy $\epsilon$-differential privacy. However, if $\mathcal{A}(\cdot)$ is an MCMC algorithm that returns a single sample from the posterior, it guarantees privacy \citep{dimitrakakis2014robust,wang2015privacy,foulds2016theory,dimitrakakis2017differential}. Adding noise to posterior samples can also guarantee privacy~\citep{zhang2016differential}, though this set of noised samples $\{\tilde{Z}^{(s)}\}_{s=1}^S$ approximates some distribution $\tilde{P}_\mathcal{M}(Z\,|\,Y)$ that depends on $\epsilon$ and is different than the exact posterior (but close, in some sense, and equal when $\epsilon \rightarrow 0$).  For specific models, we can also noise the transition kernel of the MCMC algorithm to construct a Markov chain whose stationary distribution is again not the exact posterior, but something close that guarantees privacy \citep{foulds2016theory}.  We can also take an analogous approach to privatize variational inference, wherein we add noise to the sufficient statistics computed in each iteration \citep{park2016private}.\looseness=-1

\textbf{Locally private Bayesian inference.} We first formalize the general objective of Bayesian inference under local privacy.  Given a generative model $\mathcal{M}$ for non-privatized data $Y$ and latent variables $Z$ with joint distribution $P_{\mathcal{M}}(Y, Z)$, we further assume a randomized response method $\mathcal{R}(\cdot)$ that generates privatized data sets: $\tilde{Y} \sim P_{\mathcal{R}}(\tilde{Y}\,|\,Y)$. The goal is then to approximate the \textit{locally private posterior}:~\looseness=-1
\begin{align}
  P_{\mathcal{M,R}}(Z\,|\,\tilde{Y}) &=
  \mathbbm{E}_{P_{\mathcal{M,R}}(Y\,|\,\tilde{Y})}\left[{P_{\mathcal{M}}(Z\,|\,Y)}
    \right] \notag\\
  \label{eq:objective}
  &= \int {P_{\mathcal{M}}(Z\,|\,Y)} \,
  P_{\mathcal{M,R}}(Y\,|\,\tilde{Y}) \, dY.
  \end{align}
This distribution correctly characterizes our uncertainty about the latent variables $Z$, conditioned on all of our observations and assumptions---i.e., the privatized data $\tilde{Y}$, the model $\mathcal{M}$, and the randomized response method $\mathcal{R}$. The expansion in equation~\ref{eq:objective} shows that this posterior implicitly treats the non-privatized data set $Y$ as a latent variable and marginalizes over it using the mixing distribution $P_{\mathcal{M,R}}(Y\,|\,\tilde{Y})$ which is itself a posterior that characterizes our uncertainty about $Y$ given all we observe. The central point here is that if we can generate samples from $P_{\mathcal{M,R}}(Y\,|\,\tilde{Y})$, we can use them to approximate the expectation in equation~\ref{eq:objective}, assuming that we already have a method for approximating the non-private posterior $P_{\mathcal{M}}(Z\,|\,Y)$.  In the context of MCMC, iteratively re-sampling values of the non-privatized data from their complete conditional---i.e., $Y^{(s)} \sim P_{\mathcal{M, R}}(Y \,|\, Z^{(s -1)}, \tilde{Y})$---and then re-sampling values of the latent variables---i.e., $Z^{(s)} \sim P_{\mathcal{M}}(Z\,|\,Y^{(s)})$---constitutes a Markov chain whose stationary distribution is $P_{\mathcal{M, R}}(Z, Y\,|\,\tilde{Y})$.  In scenarios where we already have derivations and implementations for sampling from $P_{\mathcal{M}}(Z\,|\,Y)$, we need only be able to sample efficiently from $P_{\mathcal{M, R}}(Y \,|\, Z, \tilde{Y})$ in order to obtain a locally private Bayesian inference algorithm; whether we can do this efficiently depends on our choice of $\mathcal{M}$ and $\mathcal{R}$.

We note that the objective of Bayesian inference under local privacy, as defined in equation~\ref{eq:objective}, is similar to that of \citet{williams2010probabilistic}, who identify their key barrier to inference as being unable to analytically form the marginal likelihood that links the privatized data to $Z$:~\looseness=-1
\begin{equation}
  \label{eq:margll}
  P_{\mathcal{M, R}}(\tilde{Y}\,|\,Z) = \int
  P_{\mathcal{R}}(\tilde{Y}\,|\,Y) \,
  P_{\mathcal{M}}(Y\,|\,Z) \, dY.
\end{equation}
In the next sections, we show that if $\mathcal{M}$ is a Poisson factorization model and $\mathcal{R}$ is the geometric mechanism, then we can form an augmented version of this marginal likelihood analytically and derive an MCMC algorithm that samples efficiently from the posterior in equation~\ref{eq:objective}.~\looseness=-1

%%%%%%%%%%%%%%%%%%%%%%%%%%%%%%%%%%%%%%%%%%%
\section{Locally private Poisson factorization}
\label{sec:poisson_factorization}

In this section, we describe a particular model $\mathcal{M}$---i.e., Poisson factorization---and randomized response method $\mathcal{R}$---i.e., the geometric mechanism---that are each natural choices for count data. We prove two theorems about $\mathcal{R}$: 1) that it is a mechanism for LPLP and 2) that it can be re-interpreted in terms of the Skellam distribution~\citep{skellam1946frequency}. We rely on the second theorem to show that our choices of $\mathcal{M}$ and $\mathcal{R}$ combine to yield a novel generative process for privatized count data which we exploit, in the next section, to derive efficient Bayesian inference.~\looseness=-1

\begin{figure*}[t]
\centering
\includegraphics[width=\linewidth]{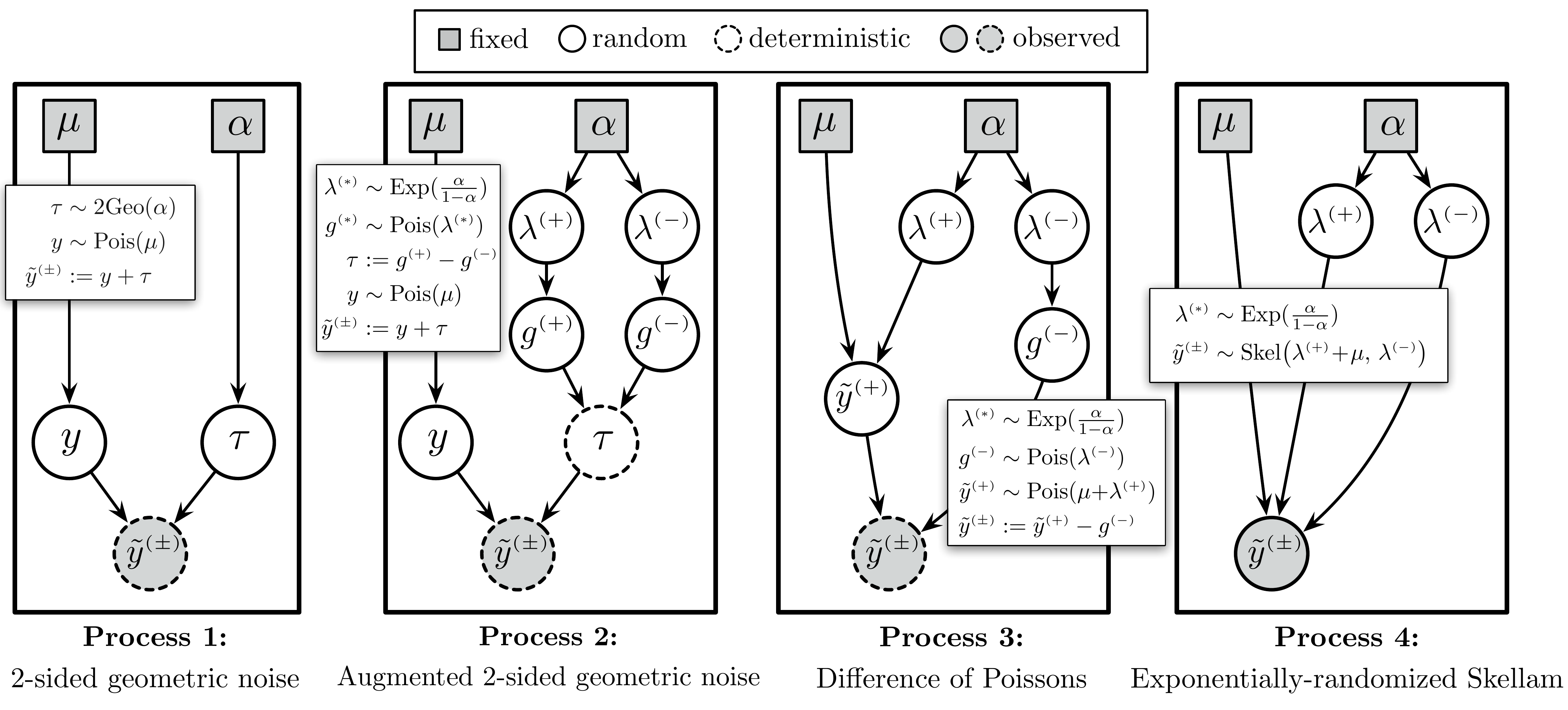}\vspace{-1em}
 \caption{\label{fig:four_ways}Four generative processes that yield the same marginal distributions $P(\tilde{y}^{(\pm)} \given \mu, \alpha)$. Process 1 generates $y^{\mathsmaller{(\pm)}}$ as the sum of an independent Poisson and two-sided geometric random variable. Process 2 augments the two-sided geometric random variable as the difference of two Poisson random variables with exponentially-randomized rates. Process 3 represents the sum of $y$ and the additive geometric random variable $g^{\mathsmaller{(+)}}$ as a single Poisson random variable $\tilde{y}^{\mathsmaller{(+)}}$. Process 4 marginalizes out the Poisson random variables to yield a generative process for $\tilde{y}^{\mathsmaller{(\pm)}}$ as a Skellam random variable with exponentially-randomized rates.~\looseness=-1}
 \vspace{-1em}
\end{figure*}

\textbf{$\mathcal{M}$: Poisson factorization.} We assume that $Y$ is a data set consisting of counts, each of which $\ys \tin \mathbbm{Z}_+$ is an independent Poisson random variable $\ys \tsim \textrm{Pois}(\mus)$ with rate $\mus$ that is defined to be a deterministic function of shared model parameters $Z$. The subscript $\subs$ is a multi-index. In Poisson \emph{matrix} factorization $\subs \tequiv (\subs_1, \subs_2)$. However this formalism includes tensor factorization $\subs \tequiv (\subs_1,\dots,\subs_M)$ and/or multiview models in which the multi-index across counts may differ in the number of indices. This class of models includes many widely used models in social science, as described in section~\ref{sec:intro}. In section~\ref{sec:case_studies}, we provide case studies using two different Poisson factorization models---i.e., mixed-membership stochastic block model for social networks~\citep{ball2011efficient,gopalan2013efficient,zhou2015infinite} and latent Dirichlet allocation~\citep{blei2003latent}---which we describe in more detail. We selected these models for purposes of exposition---they are simple and widely used. Although both are instances of Poisson \emph{matrix} factorization, the method we present in this paper applies to any Poisson factorization model.~\looseness=-1

\textbf{$\mathcal{R}$: Geometric mechanism.} The two most commonly used randomized response mechanisms in the privacy toolbox---the Gaussian and Laplace mechanisms---privatize observations by adding noise drawn from continuous real-valued distributions; they are thus unnatural choices for count  data. \citet{ghosh2012universally} introduced the geometric mechanism, which can be viewed as the discrete analog to the Laplace mechanism, and involves adding integer-valued noise $\tau \in \mathbb{Z}$ drawn from the two-sided geometric distribution $\tau \sim \textrm{2Geo}(\alpha)$; the PMF of this distribution is:\looseness=-1
 \begin{equation}
   \label{eq:2side}
   \textrm{2Geo}(\tau; \alpha) =
   \frac{1-\alpha}{1+\alpha}\,\alpha^{|\tau|}.
   \end{equation}
\begin{theorem}
\label{theorem:llp}
\textup{(Proof in appendix)} 
Let randomized response method $\mathcal{R}(\cdot)$ be the geometric mechanism with parameter $\alpha$. Then for any positive integer $N$, and any pair of observations $y, y' \in \mathcal{Y}$ such that $\|y - y'\|_1 \leq N$, $\mathcal{R}(\cdot)$ satisfies
\begin{equation}
    P\left(\mathcal{R}(y) \in \mathcal{S}\right) \leq e^\epsilon
      P\left(\mathcal{R}(y') \in \mathcal{S}\right),
        \end{equation}
for all subsets $\mathcal{S}$ in the range of $\mathcal{R}(\cdot)$,
where
\begin{equation}
  \epsilon = N\ln{\Big(\frac{1}{\alpha}\Big)}.
\end{equation}
Therefore, for any $N$, the geometric mechanism with parameter $\alpha$ is an $(N,
\epsilon)$-private randomized response method with $\epsilon = N
\ln{(\frac{1}{\alpha})}$. If a data analysis algorithm sees only the observations' $(N, \epsilon)$-private responses, then the data analysis satisfies $(N, \epsilon)$-limited precision local privacy.
\end{theorem}
% A two-sided geometric random variable can be generated as the difference of two i.i.d.~geometric random variables---i.e., $\tau := g^{\mathsmaller{(+)}} - g^{\mathsmaller{(-)}}$ where $g^{\mathsmaller{(*)}} \sim \textrm{Geom}(1 \tm \alpha)$.
\begin{theorem}
  \label{theorem:geo_skellam}
\textup{(Proof in appendix)} A two-sided geometric random variable
$\tau \sim \textrm{2Geo}(\alpha)$ can be generated as follows:
\begin{equation}
    \label{eq:skell}
    \tau  \sim \textrm{Skel}(\lambda^{\mathsmaller{(+)}}, \lambda^{\mathsmaller{(-)}}), \hspace{1em}
    \lambda^{\mathsmaller{(*)}}\sim
    \textrm{Exp}(\tfrac{\alpha}{1-\alpha}),
\end{equation}
where $\textrm{Exp}(\cdot)$ is the exponential distribution and the Skellam distribution is the marginal distribution over the difference $\tau := g^{(+)} \tm g^{(-)}$ of two independent Poisson random variables $g^{(*)} \sim \text{Pois}(\lambda^{(*)})$ for $* \in \{+,-\}$.\looseness=-1
\end{theorem}

\textbf{Combining $\mathcal{M}$ and $\mathcal{R}$.} We assume each true count is generated by $\mathcal{M}$---i.e., $\ys \tsim \textrm{Pois}(\mus)$---then privatized by $\mathcal{R}$:~\looseness=-1
\begin{equation}
    \label{eq:add_noise}
      \yspm := \ys + \tau_{\subs}, \hspace{0.75em}
      \tau_{\subs} \sim \textrm{2Geo}(\alpha).
      \end{equation}
where $\yspm$ is the privatized observation which we superscript with $(\pm)$ to denote that it may be non-negative or negative since the additive noise $\tau_\subs \tin \mathbbm{Z}$ may be negative.

Via theorem~\ref{theorem:geo_skellam}, we can express the generative process for $\yspm$ in four equivalent ways, shown in figure~\ref{fig:four_ways}, each of which provides a unique and necessary insight. Process 1 is a graphical representation of the generative process as defined thus far. Process 2 represents the two-sided geometric noise in terms of a pair of Poisson random variables with exponentially distributed rates; in so doing, it reveals the auxiliary variables that facilitate inference. Process 3 represents the sum of the true count and the positive component of the noise as a single Poisson random variable $\ysp \teq \ys \tp \gp$. Process 4 marginalizes out the remaining Poisson random variables to obtain a marginal representation $\yspm$ as a Skellam random variable with exponentially-randomized rates:~\looseness=-1
\begin{equation}
    \label{eq:skell2}
    \yspm  \tsim \textrm{Skel}\left(\lambdap \tp \mus, \lambdam\right), \hspace{0.75em}
    \lambda_{\subs}^{\mathsmaller{(*)}} \tsim
    \textrm{Exp}(\tfrac{\alpha}{1-\alpha}).
\end{equation}
The derivation of processes 2--4 relies on properties of the two-sided geometric, Skellam, and Poisson distributions which we provide in the appendix.~\looseness=-1

%%%%%%%%%%%%%%%%%%%%%%%%%%%%%%%%%%%%%%%%%%%

\section{MCMC algorithm}
\label{sec:MCMC}
Upon observing a privatized data set $\tilde{Y}^{(\pm)}$, the goal of a Bayesian agent is to approximate the locally private posterior. As explained in section~\ref{sec:bg}, to do so with MCMC, we need only be able to sample the true data $\ys$ as a latent variable from its complete conditional $P_{\mathcal{M},\mathcal{R}}(\ys \given \yspm, \mus, -)$. By assuming that the privatized observations $\yspm$ are conditionally independent Skellam random variables, as in equation~\ref{eq:skell2}, and we may exploit the following general theorem that relates the Skellam to the Bessel~\citep{yuan2000bessel} distribution.~\looseness=-1

\begin{theorem}
  \label{theorem:rel}
\textup{(Proof in appendix)} Consider two Poisson random variables
$y_1 \sim \textrm{Pois}(\lambda^{\mathsmaller{(+)}})$ and $y_2 \sim
\textrm{Pois}(\lambda^{\mathsmaller{(-)}})$. Their minimum $m := \textrm{min}\{y_1, y_2\}$ and their difference $\tau := y_1 - y_2$ are deterministic functions of $y_1$ and $y_2$.  However, if not conditioned on $y_1$ and $y_2$, the random variables $m$ and $\tau$ can be marginally generated as follows:\looseness=-1
\begin{equation}
  \tau \sim \textrm{Skel}(\lambda^{\mathsmaller{(+)}}, \lambda^{\mathsmaller{(-)}}), \hspace{0.1em}
  m \sim \textrm{Bes}\left(|\tau|,
  2\sqrt{2\lambda^{\mathsmaller{(+)}}\lambda^{\mathsmaller{(-)}}}\right).\!
  \end{equation}
\end{theorem}
Theorem~\ref{theorem:rel} means that we can generate two independent Poisson random variables by first generating their difference $\tau$ and then their minimum $m$.  Because $\tau = y_1 - y_2$, if $\tau$ is positive, then $y_2$ must be the minimum and thus $y_1 = \tau-m$.
In practice, this means that if we only get to observe the difference of two Poisson-distributed counts, we can still ``recover'' the counts by sampling a Bessel auxiliary variable.

Assuming that $\yspm \sim \text{Skel}(\lambdap \tp \mus,\lambdam)$ via theorem~\ref{theorem:geo_skellam}, we first sample an auxiliary Bessel random variable $m_{\subs}$:
\begin{align}
  \label{eq:first}
  \compcond{m_{\subs}} &\sim
  \text{Bes}\left(|\yspm|,\, 2\sqrt{(\lambdap
    \tp \mus)\lambdam}\right).
  \end{align}
\citet{yuan2000bessel} give details of the Bessel distribution, which can be sampled efficiently~\citep{devroye2002simulating}.
% \footnote{We will release our implementation of Bessel sampling after review. We are unaware of any other open-source implementations.}

By theorem~\ref{theorem:rel}, $m_{\subs}$ represents the minimum of two latent Poisson random variables whose difference equals the observed $\yspm$; these two latent counts are given explictly in process 3 of figure~\ref{fig:four_ways}---i.e., $\yspm := \ysp - \gm$ and thus $m_{\subs} = \min\{\ysp, \gm\}$. Given a sample of $m_{\subs}$ and the observed value of $\yspm$, we can then compute $\ysp$, $\gm$:~\looseness=-1
\begin{align}
  \ysp &:= m_{\subs}, \hspace{0.5em} \gm :=
  \ysp - \yspm \hspace{0.5em }\text{if
  } \yspm \leq 0\\
\nonumber \gm &:= m_{\subs},\hspace{0.5em} \ysp := \gm +
  \yspm\hspace{0.5em} \textrm{otherwise}.
  \end{align}
Because $\ysp \teq \ys + \gp$ is itself the sum of two independent Poisson random variables, we can then sample $\ys$ from its conditional posterior, which is a binomial distribution:\looseness=-1
\begin{align}
    \label{eq:binom}
    \compcond{\ys} &\sim \textrm{Binom}\left(\ysp,
             {\tfrac{\mus}{\mus+ \lambdap}}\right).
             \end{align}
Equations~\ref{eq:first} through~\ref{eq:binom} sample the true underlying data $\ys$ from $P_{\mathcal{M, R}}(\ys \,|\, \yspm, \mus, \boldsymbol{\lambda}_\subs)$. We may also re-sample the auxiliary variables $\lambdap,\lambdam$ from their complete conditional, which is a gamma distribution, by conjugacy:
\begin{align}
  \label{eq:last}
  \compcond{\lambda_\subs^{\mathsmaller{(*)}}} &\sim \Gamma\left(1 + g_\subs^{\mathsmaller{(*)}},
  \mathsmaller{\frac{\alpha}{1 - \alpha}} + 1\right).
  \end{align}
Iteratively re-sampling $\ys$ and $\boldsymbol{\lambda}_\subs$ constitutes a chain whose stationary distribution over $\ys$ is $P_{\mathcal{M},\mathcal{R}}(\ys \given \yspm, \mus)$, as desired. Conditioned on a sample of the underlying data set $Y$, we then re-sample the latent variables $Z$ (that define the rates $\mus$) from their complete conditionals, which match those in standard non-private Poisson factorization. Equations~\ref{eq:first}--\ref{eq:last} along with non-private complete conditionals for $Z$ thus define a privacy-preserving MCMC algorithm that is asymptotically guaranteed to sample from the locally private posterior $P_{\mathcal{M}, \mathcal{R}}(Z \,|\, \tilde{Y}^{(\pm)})$ for any Poisson factorization model.~\looseness=-1

\section{Case studies}
\label{sec:case_studies}
We present two case studies applying the proposed method to 1) overlapping community detection in social networks and 2) topic modeling for text corpora. In each, we formulate natural local-privacy guarantees, ground them in examples, and demonstrate our method on real and synthetic data.~\looseness=-1
 
\textbf{Enron corpus data.} For our real data experiments, we obtained count matrices derived from the Enron email corpus~\cite{klimt2004enron}. For the community detection case study, we obtained a $V \ttimes V$ adjacency matrix $Y$ where $y_{ij}$ is the number of emails sent from actor $i$ to actor $j$.  We included an actor if they sent at least one email and sent or received at least one hundred emails, yielding $V\teq161$ actors. When an email included multiple recipients, we incremented the corresponding counts by one. For the topic modeling case study, we randomly selected $D \teq 10,000$ emails with at least 50 word tokens. We limit the vocabulary to $V \teq10,000$ word types, selecting only the most frequent word types with document frequency less than 0.3. In both case studies, we privatize the data using the geometric mechanism under varying degrees of privacy and examine each method's ability to reconstruct the true underlying data.\looseness=-1

\textbf{Reference methods.} We compare the performance of our method to two references methods: 1) the non-private approach---i.e., standard Poisson factorization fit to the true underlying data, and 2) the \naive approach---i.e., standard Poisson factorization fit to the privatized data, as if it were the true data. The \naive approach must first truncate any negative counts $\yspm \!<\! 0$ to zero and thus implicitly uses the \emph{truncated} geometric mechanism~\citep{ghosh2012universally}.~\looseness=-1

\textbf{Performance measures.} All methods generate a set of $S$ samples of the latent variables using MCMC. We use these samples to approximate the posterior expectation of $\ys$:~\looseness=-1
\begin{equation}
  \label{eq:proxy}
  \hat{\mu}_\subs = \frac{1}{S}\sum_{s=1}^S \mu^{(s)}_\subs \approx \mathbbm{E}_{P_{\mathcal{M}, \mathcal{R}}(Z\,|\,\tilde{Y}^{(\pm)})}\left[\mus\right].
\end{equation}
We then calculate the mean absolute error (MAE) of $\hat{\mu}_\subs$ with respect to the true data $\ys$. In the topic modeling case study, we also compare the quality of each method's inferred latent representation using two different standard metrics.
\looseness=-1

\begin{figure*}[t]
  \centering
  \includegraphics[width=0.33\linewidth]{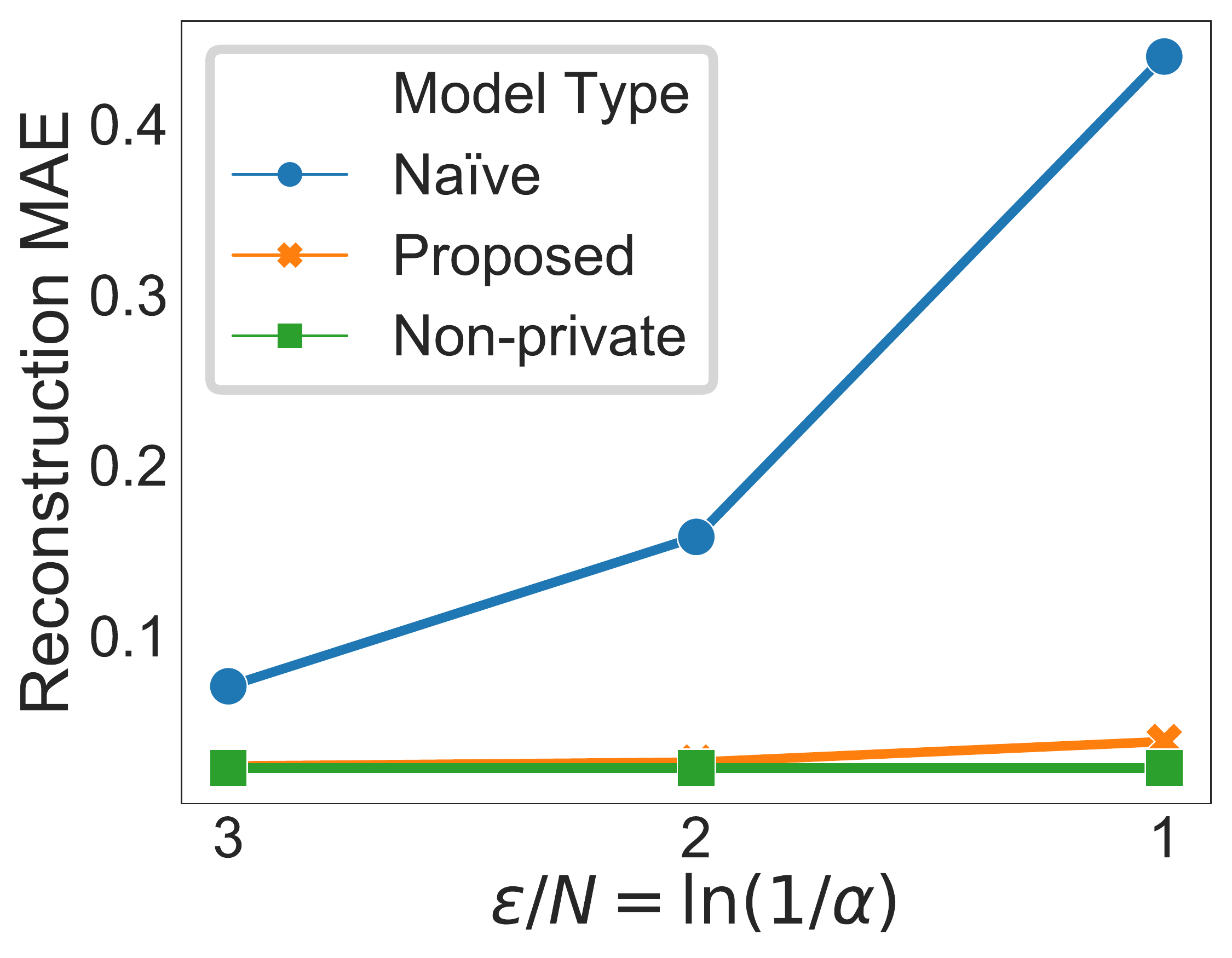}
  \includegraphics[width=0.33\linewidth]{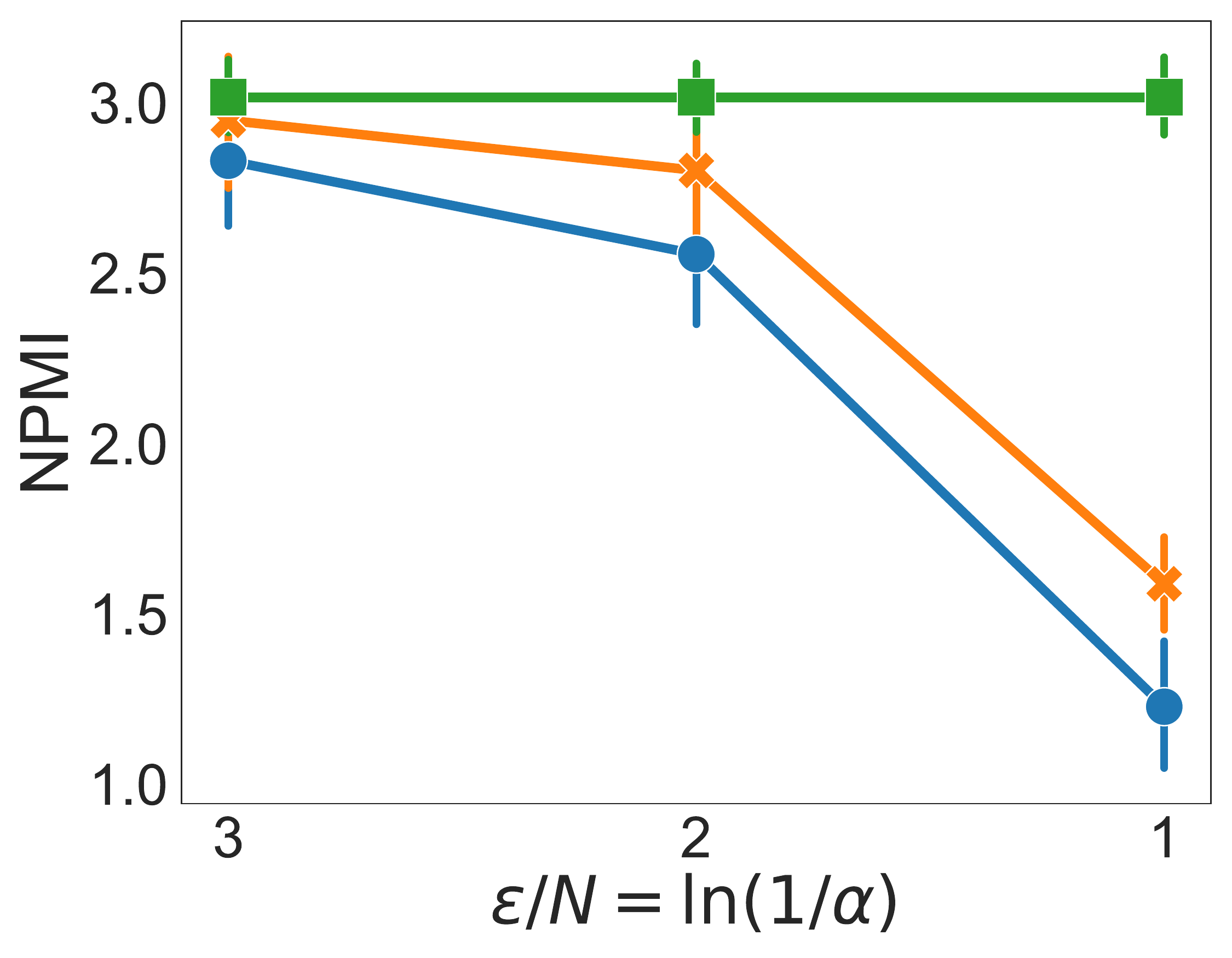}
  \includegraphics[width=0.33\linewidth]{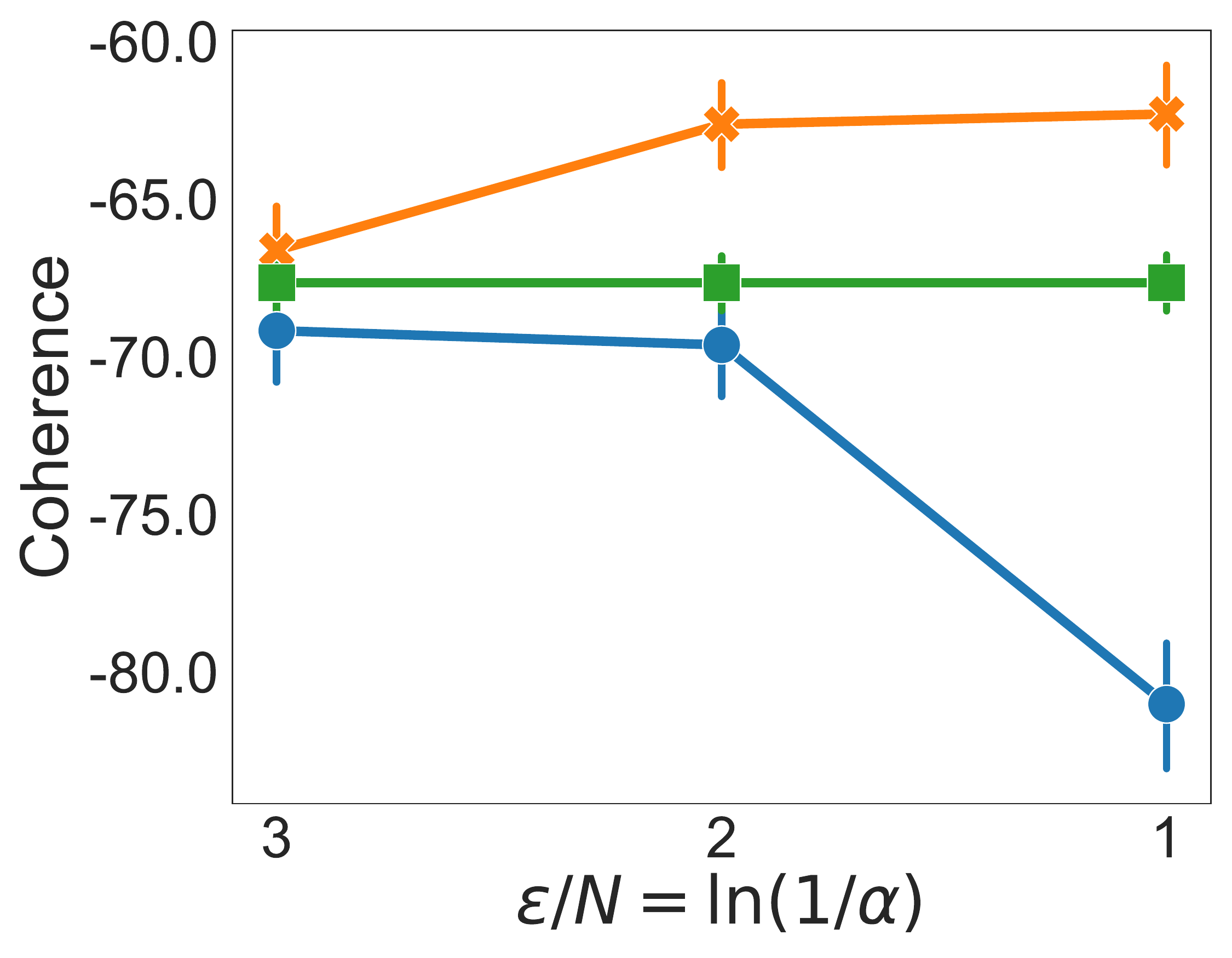}\vspace{-0.75em}
  \caption{The proposed approach obtains higher quality topics and lower reconstruction error than the \naive approach. Each plot compares the proposed, \naive, and non-private approaches for three increasing levels of noise (privacy) on the Enron corpus data; the non-private values are constant across privacy levels. The \textit{left} plot measures performance using mean absolute error (MAE) of the estimated Poisson rates $\hat{\mu}_{dv}$ against the true underlying data $y_{dv}$; \textit{lower is better}. The \textit{center} and \textit{right} plots measure topic quality/interpretability using normalized pointwise mutual information (NPMI) and coherence, respectively, where \textit{higher is better}. When measuring using coherence (\textit{right}), the proposed approach obtains higher quality topics than the non-private approach. \label{fig:topic-interpretability}}\vspace{-1em}
  \end{figure*}

\subsection{Case study 1: Topic modeling}
\label{sec:casetopic}

Topic models~\citep{blei2003latent} are widely used in the social sciences~\citep{ramage2009topic,grimmer2013text,mohr2013poetics,roberts2013structural} for characterizing the high-level thematic structure of text corpora via latent ``topics''---i.e., probability distributions over vocabulary items. In many settings, documents contain sensitive information (e.g., emails, survey responses) and individuals may be unwilling to share their data without formal privacy guarantees, such as those provided by differential privacy.

\textbf{Limited-precision local privacy.} In this scenario, a data set $Y$ is a $D \ttimes V$ count matrix, each element of which $y_{dv} \tin \mathbbm{Z}_+$ represents the count of word type $v \tin [V]$ in document $d \tin [D]$. It is natural to consider each document $\boldsymbol{y}_d \tequiv (y_{d1},\dots,y_{dV})$ to be a single ``observation'' we seek to privatize. Under LPLP, the precision level $N$ determines the neighborhood of documents within $\epsilon$-level local privacy applies. For instance, if $N\teq 4$, then two emails---one that contained four instances of a vulgar curse word and the same one that did not $\|\boldsymbol{y}_d - \boldsymbol{y}'_d\|_1=4$---would be rendered indistinguishable after privatization, assuming small $\epsilon$.~\looseness=-1

% In this case, if $y_{dv} \leq N$, an adversary would be unable to tell from $\tilde{y}_{dv}^{(\pm)}$ whether $v$ occurred in $d$, provided $\epsilon$ is sufficiently small. However, a more natural interpretation would be to assume that a single observation is an entire document---i.e., $\boldsymbol{y}_d = (y_{d1}, \ldots, y_{dV})$. In this case, if the $\ell_1$ norm of the difference between two documents is $N$ or less, then their privatized versions will be indistinguishable, provided $\epsilon$ is sufficiently small. For example, if $N = 4$, then the privatized version of email that includes the sentence ``I hate my boss'' will be indistinguishable from that of an email without the sentence. We note that it is also natural to consider heterogeneous document-specific precision levels---i.e., $N_d$ leading to $\epsilon_d = N_d \ln{\left( \frac{1}{\alpha_d}\right)}$---to enable the author of document $d$ to choose how much to noise this document before sharing it. For example, if an author wanted to make sure that an adversary would be unable to tell that she wrote ``surprise party'' five times in an email, she would first set $N_d := 5 \cdot 2 = 10$. Then, to achieve $\epsilon_d = 1$, she would set $\tilde{y}_{dv}^{(\pm)} := y_{dv} + \tau_{dv}$, where $\tau_{dv} \sim \textrm{2Geo}(\alpha_d)$ and $\alpha_d = \exp{(-\frac{\epsilon_d}{N_d})} \approx 0.9$.\looseness=-1

\textbf{Poisson factorization}. Gamma--Poisson matrix factorization is commonly used for topic modeling. In this model, $Y$ is a $D \ttimes V$ count matrix. We assume each element is drawn $y_{dv} \tsim \Pois(\mu_{dv})$ where $\mu_{dv} = \sum_{k=1}^K \theta_{dk}\, \phi_{kv}$. The factor $\theta_{dk}$ represents how much topic $k$ is used in document $d$, while the factor $\phi_{kv}$ represents how much word type $v$ is used in topic $k$. The set of latent variables is thus $Z = \{\Theta, \Phi\}$, where $\Theta$ and $\Phi$ are $D \ttimes K$ and $K \ttimes V$ non-negative, real-valued matrices, respectively. It is standard to assume independent gamma priors over the factors---i.e., $\theta_{dk}, \phi_{kv} \tsim \Gamma(a_0, b_0)$, where we set the shape and rate hyperparameters to $a_0 \teq 0.1$ and $b_0 \teq 1$, respectively.\looseness=-1

\textbf{Enron corpus experiments}. In these experiments, we use the document-by-term count matrix $Y$ derived from the Enron email corpus. We consider three privacy levels $\epsilon / N \in \{3, 2, 1\}$ specified by the ratio of the privacy budget $\epsilon$ to the precision $N$. For each privacy level, we obtain five different privatized matrices, each by adding two-sided geometric noise with $\alpha \teq -\exp (\epsilon / N)$ independently to each element. We fit both privacy-preserving models---i.e., ours and the \naive approach---to all five privatized matrices for each privacy level. We also fit the non-private approach five independent times to the true matrix. For all models we used $K\teq 50 $ topics. For every model and matrix, we perform 7,500 MCMC iterations, saving every $100^{\textrm{th}}$ sample of the latent variables $\Phi^{(s)}, \Theta^{(s)}$ after the first 2,500. We use the fifty saved samples to compute $\hat{\mu}_{dv}=\frac{1}{S}\sum_{s=1}^S \sum_{k=1}^K \theta_{dk}^{(s)} \phi_{kv}^{(s)}$.~\looseness=-1

\textbf{Results}. We find that our approach obtains both lower reconstruction error and higher quality topics than the \naive approach. For each model and matrix, we compute the mean absolute error (MAE) of the reconstruction rates $\hat{\mu}_{dv}$ with respect to the true underlying matrix $Y$. These results are visualized in the left subplot of figure~\ref{fig:topic-interpretability} where we see that the proposed approach reconstructs the true data with nearly as low error as non-private inference (that fits to the true data) while the \naive approach has high error which increases dramatically as the noise (i.e., privacy) increases.

To evaluate topic quality, we use two standard metrics---i.e., normalized pointwise mutual information (NPMI)~\citep{lau2014machine} and topic coherence~\citep{mimno2011optimizing}--- applied to the 10 words with maximum weight for each sampled topic vector $\boldsymbol{\phi}^{(s)}_k$, using the true data as the reference corpus. For each method and privacy level, we average these values across samples.  The center and right subplots of figure~\ref{fig:topic-interpretability} visualize the NPMI and coherence results, respectively. The proposed approach obtains higher quality topics than the \naive approach, as measured by both metrics. As measured by coherence, the proposed approach even obtains higher quality topics than the non-private approach.~\looseness=-1

\textbf{Synthetic data experiments.} We also find that the proposed approach is more effective than the \naive approach at recovering the ground-truth topics $\Phi^*$ from synthetically-generated data. For space reasons, we include these results along with a description of the experiments in the appendix.

\begin{figure*}[t]
    \begin{subfigure}[t]{0.49\linewidth}
      \centering
      \includegraphics[width=\linewidth]{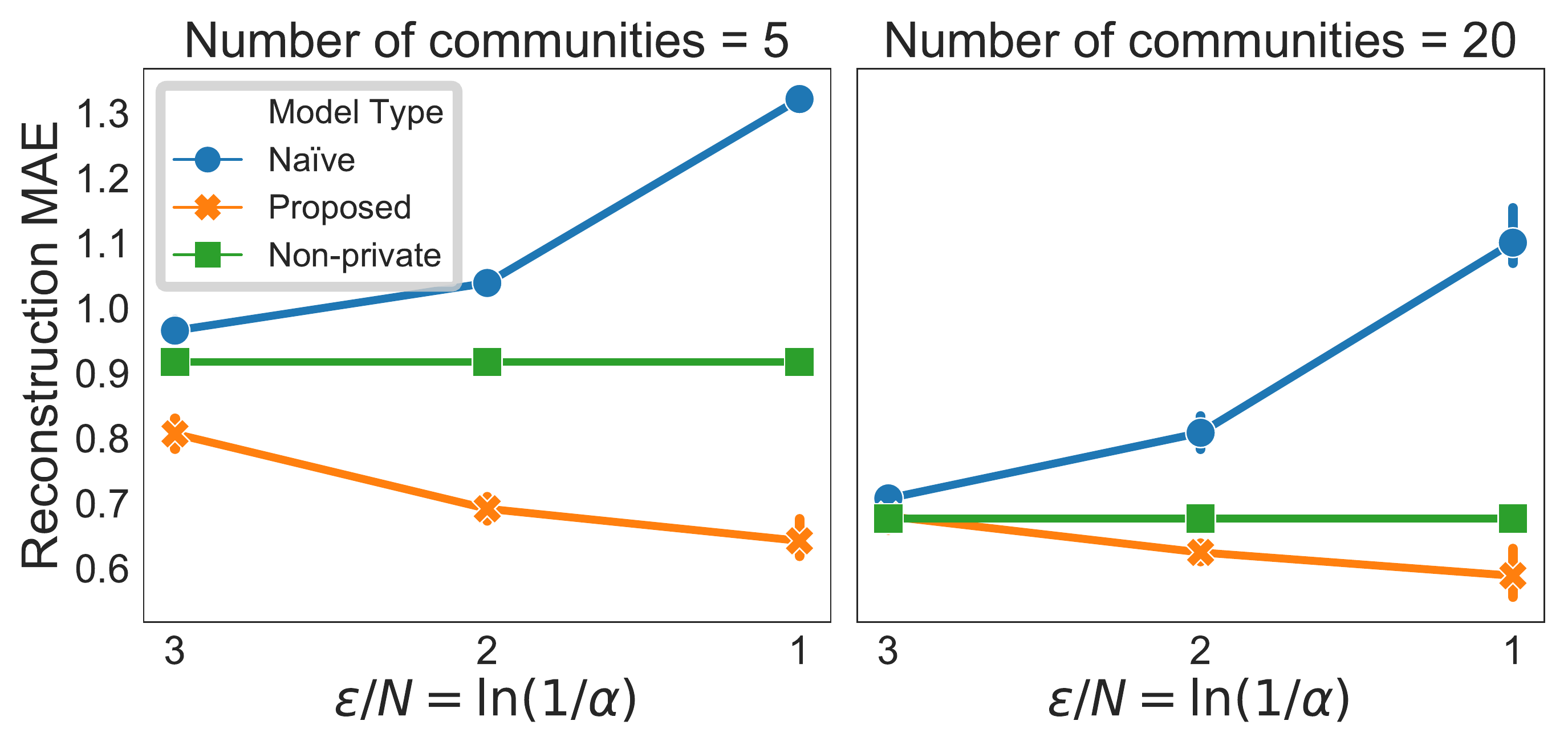}
      % \caption{}
    \end{subfigure}
    \begin{subfigure}[t]{0.49\linewidth}
      \centering
      \includegraphics[width=\linewidth]{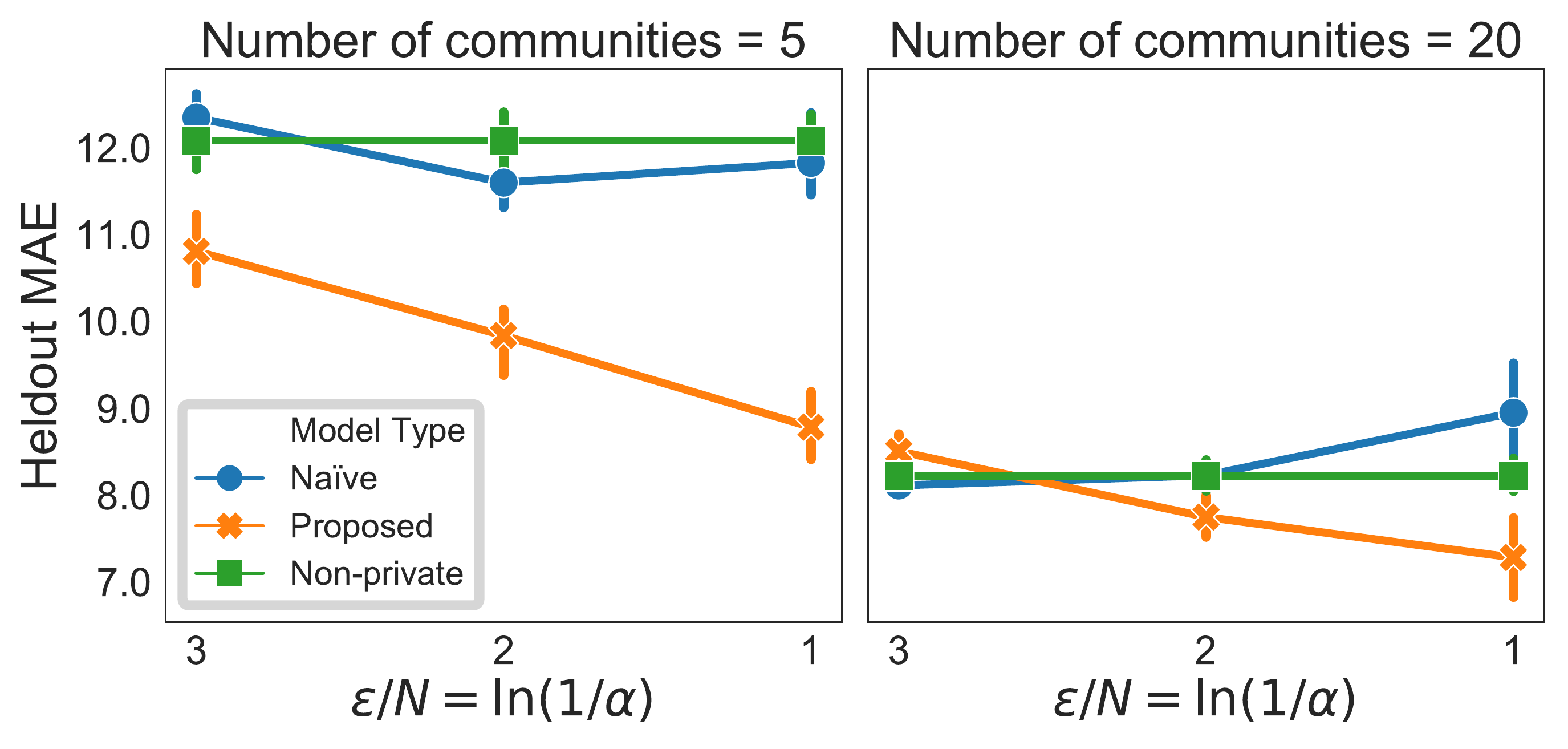}
      % \caption{}
    \end{subfigure}\vspace{-0.75em}
  \caption{The proposed approach obtains lower error on both reconstruction and heldout link prediction than the \naive and even the non-private approach. \emph{Left and center left}: The methods estimate $\hat{\mu}_{ij}$ from the fully observed matrix. \emph{Center right and right}:  The methods estimate $\hat{\mu}_{ij}$ from a partially observed matrix; MAE is computed with respect to only the heldout entries $y_{ij}$. All experiments are repeated for different numbers of communities $C \tin \{5,10,20\}$; the results are similar across all values (we include $C\teq 10$ in the appendix).~\looseness=-1 \label{fig:comm-mae}}\vspace{-1em}
\end{figure*}

\subsection{Case study 2: Overlapping community detection}

Organizations often ask: are there missing connections between employees that, if present, would significantly reduce duplication of effort? Though social scientists may be able to draw insights based on employees' interactions, sharing such data risks privacy violations. Moreover, standard anonymization procedures can be reverse-engineered adversarially and thus do not provide privacy guarantees~\citep{narayanan2009de-anonymizing}. In contrast, the formal privacy guarantees provided by differential privacy may be sufficient for employees to consent to sharing their data.\looseness=-1

\textbf{Limited-precision local privacy.} In this setting, a data set $Y$ is a $V \ttimes V$ count matrix, where each element $y_{ij} \tin \mathbbm{Z}_+$ represents the number of interactions from actor $i \tin [V]$ to actor $j \tin [V]$. It is natural to consider each count $y_{ij}$ to be a single ``observation''. Using the geometric mechanism with $\alpha = -\exp(\epsilon/N)$, if $i$ interacted with $j$ three times $y_{ij} \teq 3$ and $N \teq 3$, then an adversary would be unable to tell from $\tilde{y}_{ij}^{\mathsmaller{(\pm)}}$ whether $i$ had interacted with $j$ at all, provided $\epsilon$ is sufficiently small. Note  that if $y_{ij} \gg N$, then an adversary would be able to tell that $i$ had interacted with $j$, though not the exact number of times.\looseness=-1

\pagebreak
\textbf{Poisson factorization model.} The mixed-membership stochastic block model for learning latent overlapping community structure in social networks~\citep{ball2011efficient,gopalan2013efficient,zhou2015infinite} is a special case of Poisson factorization where $Y$ is a $V \ttimes V$ count matrix, each element of which is assumed to be drawn $y_{ij} \tsim \Pois(\mu_{ij})$ where $\mu_{ij} \teq \sum_{c=1}^C \sum_{d=1}^C \theta_{ic}\,\theta_{jd}\,\pi_{cd}$. The factors $\theta_{ic}$ and $\theta_{jd}$ represent how much actors $i$ and $j$ participate in communities $c$ and $d$, respectively, while the factor $\pi_{cd}$ represents how much actors in community $c$ interact with actors in community $d$. The set of latent variables is thus $Z \teq \{\Theta, \Pi\}$ where $\Theta$ and $\Pi$ are $V \ttimes C$ and $C \ttimes C$ non-negative, real-valued matrices, respectively. It is standard to assume independent gamma priors over the factors---i.e., $\theta_{ic}, \pi_{cd} \tsim \Gamma(a_0, b_0)$, where we set the shape and rate hyperparameters to $a_0 \teq 0.1$ and $b_0 \teq 1$, respectively.\looseness=-1

% \begin{figure*}[t]
%   \label{fig:example}
%     \begin{center}
%       \includegraphics[width=\linewidth]{figs/enron_community_MAE.pdf}
%     \end{center}
%     \vspace{-2em}
%     \caption{Mean average error (MAE) between the true Enron data $Y$ and the
%     estimate $\hat{\mu}$. The dotted line is the baseline from non-private
%     inference, i.e. when no noise is added. Surprisingly, where the naive method's error increases with the level of privacy, our method improves, perhaps due to the utility of random noise in preventing overfitting.\swcomment{add explanation on x-axis corresponds to level of noise}}
%     \label{fig:comm-mae}
%   \end{figure*}
% \begin{figure}[t]
% \centering
%     \includegraphics[width=0.8\linewidth]{evaluation_figures/new_figs/enron_communities/recon_mae_5.pdf}
%     \includegraphics[width=0.8\linewidth]{evaluation_figures/new_figs/enron_communities/recon_mae_20.pdf}
%     \includegraphics[width=0.8\linewidth]{evaluation_figures/new_figs/enron_communities/heldout_mae_5.pdf}
%     \includegraphics[width=0.8\linewidth]{evaluation_figures/new_figs/enron_communities/heldout_mae_20.pdf}
% \end{figure}

\textbf{Synthetic data experiments.} We generated social networks of $V\teq20$ actors with $C\teq 5$ communities. We randomly generated the true parameters $\theta^*_{ic}, \pi^*_{cd} \tsim \Gamma(a_0, b_0)$ with $a_0\teq 0.01$ and $b_0\teq 0.5$ to encourage sparsity; doing so exaggerates the block structure in the network. We then sampled a data set $y_{ij} \sim \textrm{Pois}(\mu^*_{ij})$ and added noise $\tau_{ij} \sim \textrm{2Geo}(\alpha)$ for three increasing values of $\alpha$. In each trial, we set $N$ to the empirical mean of the data $N := \hat{\mathbbm{E}}[y_{ij}]$ and then set $\alpha := \exp(-\epsilon/N)$ for $\epsilon \in \{ 2.5, 1, 0.75 \}$. For each model, we ran 8,500 MCMC iterations, saving every $25^{\textrm{th}}$ sample after the first 1,000 and using these samples to compute $\hat{\mu_{ij}}$. In figure~\ref{fig:synth-community}, we visually compare the estimates of $\hat{\mu}_{ij}$ by our proposed method to those of the \naive and non-private approaches. The \naive approach overfits the noise, predicting high rates in sparse parts of the matrix.  In contrast, the proposed approach reproduces the sparse block structure even under high noise.~\looseness=-1

\textbf{Enron corpus experiments.} For the Enron data experiments, we follow the same experimental design outlined in the topic modeling case study; we repeat this experiment using three different numbers of communities $C \in \{5, 10, 20\}$. Each method is applied to five privatized matrices for three different privacy levels. We compute $\hat{\mu}_{ij} \teq \frac{1}{S}\sum_{s=1}^S \sum_{c=1}^C \sum_{d=1}^C \theta^{(s)}_{ic} \theta^{(s)}_{jd} \pi^{(s)}_{cd}$ from each run and measure reconstruction MAE with respect to the true underlying data $y_{ij}$. In these experiments, each method observes the entire matrix ($\tilde{Y}^{\mathsmaller{(\pm)}}$ for the privacy-preerving methods and $Y$ for the non-private method). Since missing link prediction is a common task in the networks community, we additionally run the same experiments but where a portion of the matrix is masked---specifically, we hold out all entries $\tilde{y}^{\mathsmaller{(\pm)}}_{ij}$ (or $y_{ij}$ for non-private) that involve any of the top 50 senders $i$ or recipients $j$. We then compute $\hat{\mu}_{ij}$, as before, but only for the missing entries and compare heldout MAE across methods.~\looseness=-1 

\textbf{Results.} We visualize the results for $C \tin \{5,20\}$ in figure~\ref{fig:comm-mae}. The results for $C \teq 10$ are similar and given in the appendix. When reconstructing $y_{ij}$ from observed $\tilde{y}^{\mathsmaller{(\pm)}}_{ij}$, the proposed approach achieves error lower than the \naive approach and lower error than the non-private approach (which directly observes $y_{ij}$). Similarly, when predicting missing $y_{ij}$, the proposed approach achieves the lowest error in most settings.~\looseness=-1 

\section{Discussion and future work}
The proposed privacy-preserving MCMC method for Poisson factorization improves substantially over the commonly-used \naive approach. A suprising finding is that the proposed method was also often better at predicting the true $y_{ij}$ from privatized $\tilde{y}^{\mathsmaller{(\pm)}}_{ij}$ than even the non-private approach. Similarly, the the proposed approach inferred more coherent topics. These empirical findings are in fact consistent with known connections between privacy-preserving mechanisms and regularization~\citep{chaudhuri2009privacy}. The proposed approach is able to explain  natural dispersion in the true data as coming from the randomized response mechanism; it may thus be more robust---i.e., less susceptible to inferring spurious structure---than non-private Poisson factorization. Future application of the model $\ys \sim \textrm{Skel}(\lambdap \tp \mus,\, \lambdam)$ as a robust alternative to Poisson factorization is thus motivated, as is a theoretical characterization of its regularizing properties.

\pagebreak
\bibliography{paper}
\bibliographystyle{icml2019}

\end{document}